%% file: colm2026_conference.tex
\documentclass{article} 
\usepackage[final]{colm2026_conference}

\usepackage{amsmath, amssymb, amsfonts, amsthm}
\usepackage{bbm}
\usepackage{nicefrac}
\usepackage{mathabx}

\usepackage{tabularx}
\usepackage{booktabs}
\usepackage{pifont}
\usepackage[table]{xcolor}
\definecolor{forestgreen}{RGB}{34,139,34}
\usepackage{booktabs}
\usepackage{multirow}
\usepackage{makecell}
\usepackage{subcaption}
\usepackage{graphicx}
\usepackage{float}
\usepackage{stfloats}
\usepackage{wrapfig}
\usepackage{capt-of}
\usepackage[most]{tcolorbox}
\usepackage{tcolorbox}
\tcbuselibrary{skins, breakable, hooks}
\tcbuselibrary{listings}
\newtcolorbox{ReasoningBox}[1][]{
    colback=gray!5,
    colframe=gray!50,
    fonttitle=\bfseries,
    title={Analysis},
    breakable,
    #1
}
\usepackage{algorithm}
\usepackage{algpseudocode}

\usepackage[table]{xcolor}
\usepackage{hyperref}
\usepackage{url}

\usepackage{xspace}
\usepackage{enumitem}
\setlist{leftmargin=5mm}
\usepackage{textcomp}
\usepackage{pifont}
\usepackage{tcolorbox}
\usepackage{lineno}
\usepackage{latexsym}
\newtcolorbox{mybox}[1][]{
    title=#1,
    fonttitle=\small,
    fontupper=\small,
    left=2mm,
    right=2mm,
    top=1mm,
    bottom=0mm,
}

\definecolor{darkblue}{rgb}{0, 0, 0.5}
\definecolor{darkred}{rgb}{0.72, 0.22, 0.27}
\definecolor{lightblue}{RGB}{129, 209, 241}
\definecolor{forestgreen}{RGB}{34, 139, 34}
\hypersetup{
    colorlinks=true,
    citecolor=darkblue,
    linkcolor=darkblue,
    urlcolor=darkblue,
}

\newcommand\ours{\textsc{TSRouter}\xspace}

\title{\raisebox{-0.10\height}{\includegraphics[height=0.85em]{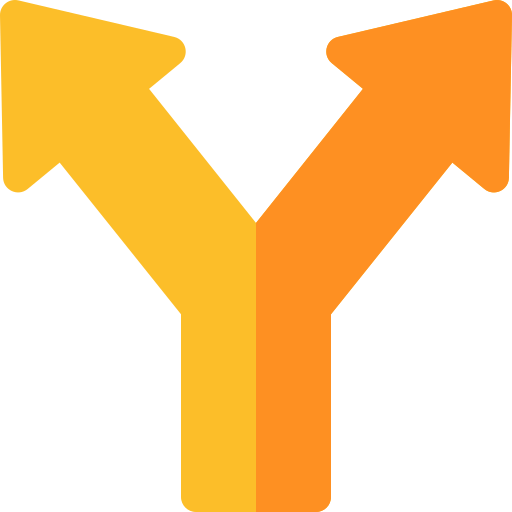}} \ours: Dynamic Modality-Model Selection for \\ Time Series Reasoning}

\author{
    \textbf{Fangxu Yu\textsuperscript{1}},\;\; \textbf{Tao Feng\textsuperscript{2}}, \;\;\textbf{Dehai Min\textsuperscript{3}}, \;\;\textbf{Lu Cheng\textsuperscript{3}}, \;\;\textbf{Ge Liu\textsuperscript{2}}, \;\;
    \textbf{Tianyi Zhou\textsuperscript{4}} \\
    \textsuperscript{1}University of Maryland, College Park, \;\;
    \textsuperscript{2}University of Illinois Urbana-Champaign,  \\
    \textsuperscript{3}University of Illinois Chicago,\;\;
    \textsuperscript{4}MBZUAI\\
}

\definecolor{barGray}{HTML}{D9D9D9} 

\begin{document}

\ifcolmsubmission
\linenumbers
\fi

\maketitle

\begin{abstract}
Time series reasoning is essential for real-world problem-solving. While both Large Language Models (LLMs) and Vision-Language Models (VLMs) can reason about time-series data, their capabilities are complementary: LLMs process time series as text sequences and thus preserve exact numerical understanding, but struggle with global patterns,
whereas VLMs efficiently capture these patterns by visualizing time series but may lose fine-grained details. Moreover, models vary significantly in task-specific expertise and inference costs. Dynamically selecting the most suitable modality and model for each query is therefore crucial, yet challenging because it requires modeling the complex interactions among tasks, queries, modalities, and models, which carry rich contextual signals.
To this end, we introduce \ours, a graph-based dynamic routing framework. \ours constructs a heterogeneous graph of task, query, modality, and model nodes to contextualize the interactions among query characteristics, modality attributes, and model capabilities. \ours formulates routing as a candidate scoring problem, where each modality-model pair is evaluated based on user-defined performance-cost preferences to select the optimal candidate. Comprehensive evaluations on 4 distinct time series reasoning tasks reveal that \ours substantially outperforms diverse baselines with 16\% to 46\% relative improvements. Furthermore, \ours demonstrates robust zero-shot plug-and-play generalization to unseen models and novel tasks and preserves high performance while reducing computational overhead through cost-aware optimization. Our code is available at \href{https://github.com/tianyi-lab/TSRouter}{https://github.com/tianyi-lab/TSRouter}.

\end{abstract}

\section{Introduction}
\label{sec:intro}
Time series plays a critical role in high-stakes domains where time series records the temporal evolution of real-world variables. Interpreting these sequential patterns is fundamental to understanding complex processes or guiding proactive decisions. For example, in education~\citep{mao2024time} , analyzing student trajectories such as engagement and attempt counts helps identify struggling learners and personalize study plans. In scientific discovery~\citep{yu2025physics}, time series such as gravitational wave recordings encode astrophysical processes that may lead to discoveries of exoplanets and black-hole mergers. Analyzing time series and conducting reasoning enable automated systems to support high-impact applications.

Recently, Large Language Models (LLMs) and Vision-Language Models (VLMs) have emerged as powerful tools for time series analysis. LLM-based methods~\citep{merrill2024language, wang2025chattime} treat time series as raw text sequences, preserving numerical precision but suffering from context length bottlenecks, high computational costs, and poor global pattern recognition. Conversely, VLM-based approaches~\citep{zhong2025time, liu2025mllm4ts} convert time series into visual plots, efficiently capturing global patterns in long sequences but losing fine-grained numerical detail due to image resolution limits. Beyond modality choice, individual models also vary in their ability to handle different types of reasoning tasks and problems. Training a single model that unifies these complementary strengths is expensive, whereas routing among off-the-shelf LLMs and VLMs offers a lightweight and scalable alternative. Therefore, dynamically routing each query to the most appropriate modality and model is crucial for fully exploiting their complementary strengths across both dimensions. However, as summarized in Table~\ref{tab: comparison}, existing routing approaches focus solely on model selection, without jointly considering modality choice. Moreover, they capture at most one type of interaction, failing to jointly model the complex relationships among tasks, queries, modalities, and models that carry rich contextual signals for routing. These limitations make them suboptimal for time series reasoning, where both modality fitness and model expertise critically determine performance.

\begin{table}[t]
    \centering
    \setlength{\tabcolsep}{12pt}
    \resizebox{\textwidth}{!}{%
    \begin{tabular}{lccccc}
        \toprule
    \multirow{2}{*}{\textbf{Method}}
      & \makecell{\textbf{Modality}}
      & \multicolumn{2}{c}{\textbf{Interaction}}
      & \multicolumn{2}{c}{\textbf{Efficiency}} \\
    \cmidrule(lr){3-4} \cmidrule(lr){5-6}
      & \textbf{Routing}
      & \textbf{Query-Query} & \textbf{Query-Model}
      & \textbf{Inference Speed} & \textbf{Memory Use} \\
        \midrule
        CausalLM~\citep{ong2024routellm}
        & \textcolor{red}{\textbf{\ding{55}}} & \textcolor{red}{\textbf{\ding{55}}} & \textcolor{red}{\textbf{\ding{55}}} & Slow & Medium \\
        GraphRouter \citep{feng2024graphrouter} & \textcolor{red}{\textbf{\ding{55}}} & \textcolor{red}{\textbf{\ding{55}}} & \textcolor{forestgreen}{\textbf{\ding{51}}} & Fast & Low \\
        Hybrid LLM \citep{ding2024hybrid} & \textcolor{red}{\textbf{\ding{55}}} & \textcolor{red}{\textbf{\ding{55}}} & \textcolor{red}{\textbf{\ding{55}}} & Medium & Medium \\
        RouterDC \citep{chen2024routerdc} & \textcolor{red}{\textbf{\ding{55}}} & \textcolor{forestgreen}{\textbf{\ding{51}}} & \textcolor{red}{\textbf{\ding{55}}} & Medium & Medium \\
        Router-R1 \citep{zhang2025router} & \textcolor{red}{\textbf{\ding{55}}} & \textcolor{red}{\textbf{\ding{55}}} & \textcolor{red}{\textbf{\ding{55}}} & Medium & High \\
        \midrule
        \rowcolor{cyan!10} \ours & \textcolor{forestgreen}{\textbf{\ding{51}}} & \textcolor{forestgreen}{\textbf{\ding{51}}} & \textcolor{forestgreen}{\textbf{\ding{51}}} & Fast & Low \\
        \bottomrule
    \end{tabular}}
    \caption{Comparison of \ours with existing routing methods. Unlike previous approaches that only select among models under a fixed modality, \ours jointly performs modality and model routing while capturing both query-query and query-model interactions with fast inference and low memory.}
        \label{tab: comparison}
\end{table}

To address these limitations, we introduce \ours, a graph-based router designed for dynamic modality and model selection. \ours constructs a heterogeneous graph to explicitly model the contextual information and interactions among tasks, queries, modalities, and models, enabling context-aware routing that generalizes to new models and tasks without retraining. 
Specifically, the graph comprises 4 distinct node types: tasks, queries, modalities, and models, with edges encoding the different relationships among them. \ours frames the model-modality routing problem as a candidate scoring task, where the optimal choice of time series modality-model relies on user-defined preferences regarding performance and cost. Once the candidates are scored, the system routes each query to the optimal candidate. To achieve zero-shot generalization to unseen tasks and models, we prompt an LLM such as GPT-5 to profile each node, outlining critical properties such as task requirements, model descriptions, and token costs. These profiles are then embedded using a pretrained text embedding model to initialize node features. Finally, a heterogeneous Graph Neural Network (GNN) aggregates cross-type neighborhood information, followed by a scorer to produce the final routing decision.

We comprehensively evaluate \ours across 4 fundamental time series reasoning tasks. Experimental results indicate that \ours greatly outperforms a wide range of baselines with 16\% - 46\% improvements. Furthermore, \ours demonstrates remarkable generalization on unseen tasks and robustly handles new models without additional fine-tuning. Finally, by explicitly optimizing for cost-aware routing, \ours establishes a superior Pareto frontier, preserving reasoning capabilities while minimizing computational overhead compared to current approaches.

\section{\ours: Modality-Model Selection for Time Series Reasoning}
\label{sec:method}

\subsection{Motivating Analysis}
We start by demonstrating the complementary benefits of different time series modalities and LLMs and VLMs for time series reasoning on the TSRBench test set (see Section~\ref{sec: exp}) and identify two observations that motivate \ours.

\begin{figure*}[t]   
	\centering
	\includegraphics[width=1.0\textwidth]{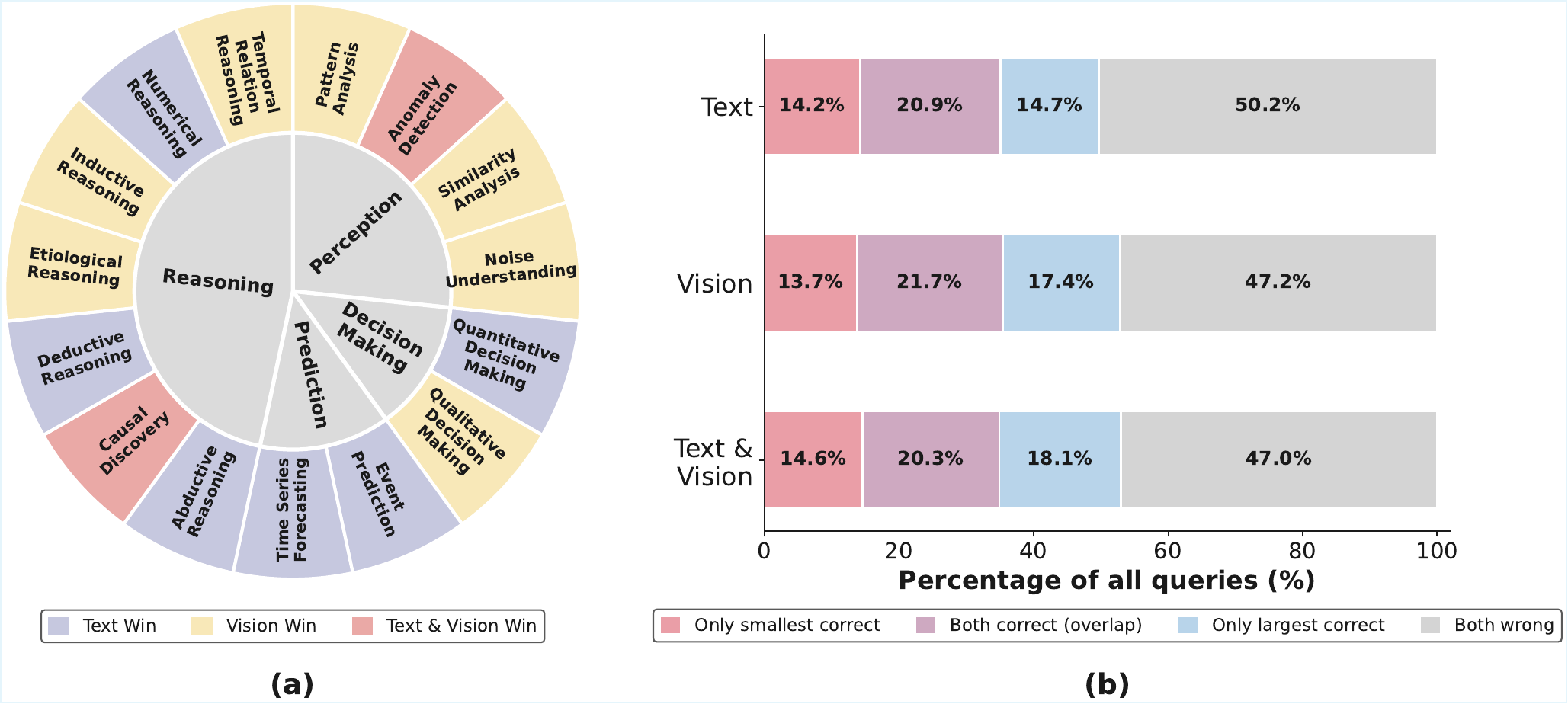}
	\caption{Motivation analysis results. (a) Distribution of best-performing modality across 15 subtasks grouped by 4 task categories. (b) Breakdown of queries correctly answered by only the smallest model, both models, and only the largest model within each modality. }
	\label{fig: preliminary}
	\vspace{-10pt}
\end{figure*}

\textbf{Different time series modalities excel on different tasks.}
Figure~\ref{fig: preliminary} (a) shows which modality achieves the highest average accuracy across models on each of the 15 subtasks, grouped by 4 task categories: Perception, Decision-Making, Prediction, and Reasoning. We observe that visual modality dominates Perception, winning 3 out of 4 subtasks, as visual plots efficiently encode the global shape and structural patterns that perception tasks rely on. In contrast, text modality excels at Prediction, where preserving exact numerical values is critical for forecasting. For Reasoning and Decision-Making, however, no single modality holds a consistent advantage. This diverse pattern confirms that different modalities complement rather than dominate others, motivating per-query modality selection rather than a fixed modality choice.
 
\textbf{Different models address varied queries.}
Figure~\ref{fig: preliminary} (b) decomposes the correctly answered queries into three disjoint groups for each modality type: those answered correctly by only the smallest model, by both models, and by only the largest model. 
Although the largest models achieve higher overall accuracy in all three categories, the overlap ratio, which is the fraction of queries answered correctly by both the smallest and largest models, is low. This indicates that larger and smaller models largely succeed on different subsets of queries: nearly half of the queries correctly answered by the large model are not the same ones correctly answered by the small model, and vice versa.
Consequently, always selecting the largest model leaves a substantial portion of queries unserved that a smaller model could have handled correctly, motivating the need for per-query routing to different models. See a qualitative analysis example in Appendix~\ref{sec: appendix motivation}.
 
These findings indicate that performance varies significantly across tasks, queries, modalities, and models, suggesting that an effective router should account for task differences, discern individual query characteristics, and jointly select the most suitable modality and model. Since these elements exhibit complex interactions that carry rich information for routing, there is a pressing need for a framework that explicitly models the interplay among them to make context-aware decisions.
\label{sec: motivation}

\subsection{Problem Formulation}
\label{sec:formulation}
Then, we formulate the time series routing problem. Let $\mathcal{Q} = \{q_1, \dots, q_N\}$ denote a set of \emph{time series queries}, where each query $q_i = (\mathbf{x}_i, u_i)$ consists of a set of time series $\mathbf{x}_i = \{x_i^{(1)}, \dots, x_i^{(K_i)}\}$ with $x_i^{(k)} \in \mathbb{R}^{T_i^{(k)}}$ and a natural language question $u_i$.
We assume access to a set of foundation models $\mathcal{L} = \{\ell_1, \dots, \ell_L\}$ containing both LLMs and VLMs, and a set of input modalities $\mathcal{M} = \{m_1, \dots, m_M\}$, where each modality specifies how the time series is presented to a model: as numerical text, as a visual chart, or as a mixture of both. LLMs can only process textual time series as inputs, and VLMs can process visual and textual inputs.
We denote the set of valid modality-model combinations as $\mathcal{C} \subseteq \mathcal{M} \times \mathcal{L}$, so that each candidate $c = (m, \ell) \in \mathcal{C}$ specifies both how the time series is presented and which model processes it.

Given a query $q_i$ and a candidate $c \in \mathcal{C}$, let $I(q_i, c) \in \{0,1\}$ denote the correctness indicator of whether $c$ produces a correct answer, and let $\text{cost}(q_i, c) \geq 0$ denote the inference cost (e.g., API pricing).
We define the \emph{effectiveness score} as a convex combination of correctness and (normalized) cost:
\begin{equation}
    \label{eq:effectiveness}
    e(q_i, c) = \alpha \cdot I(q_i, c) - (1 - \alpha) \cdot \widetilde{\text{cost}}(q_i, c),
\end{equation}
where $\widetilde{\text{cost}}$ denotes the min-max normalized cost across all candidates, and $\alpha \in [0,1]$ controls the user's preference over the performance--cost trade-off. When $\alpha = 1$, the objective reduces to pure accuracy maximization; smaller values of $\alpha$ increasingly penalize expensive candidates.
The goal of routing is to select the candidate with the highest effectiveness:
\begin{equation}
    \label{eq:objective}
    c_{i}^{*} = \arg\max_{c \in \mathcal{C}} e(q_{i}, c)
\end{equation}
where $c_i^*$ is the optimal candidate for solving query $q_i$.

\subsection{\ours Framework}
\label{sec:graph}
\begin{figure*}[t]   
	
\centering

\includegraphics[width=1.0\textwidth]{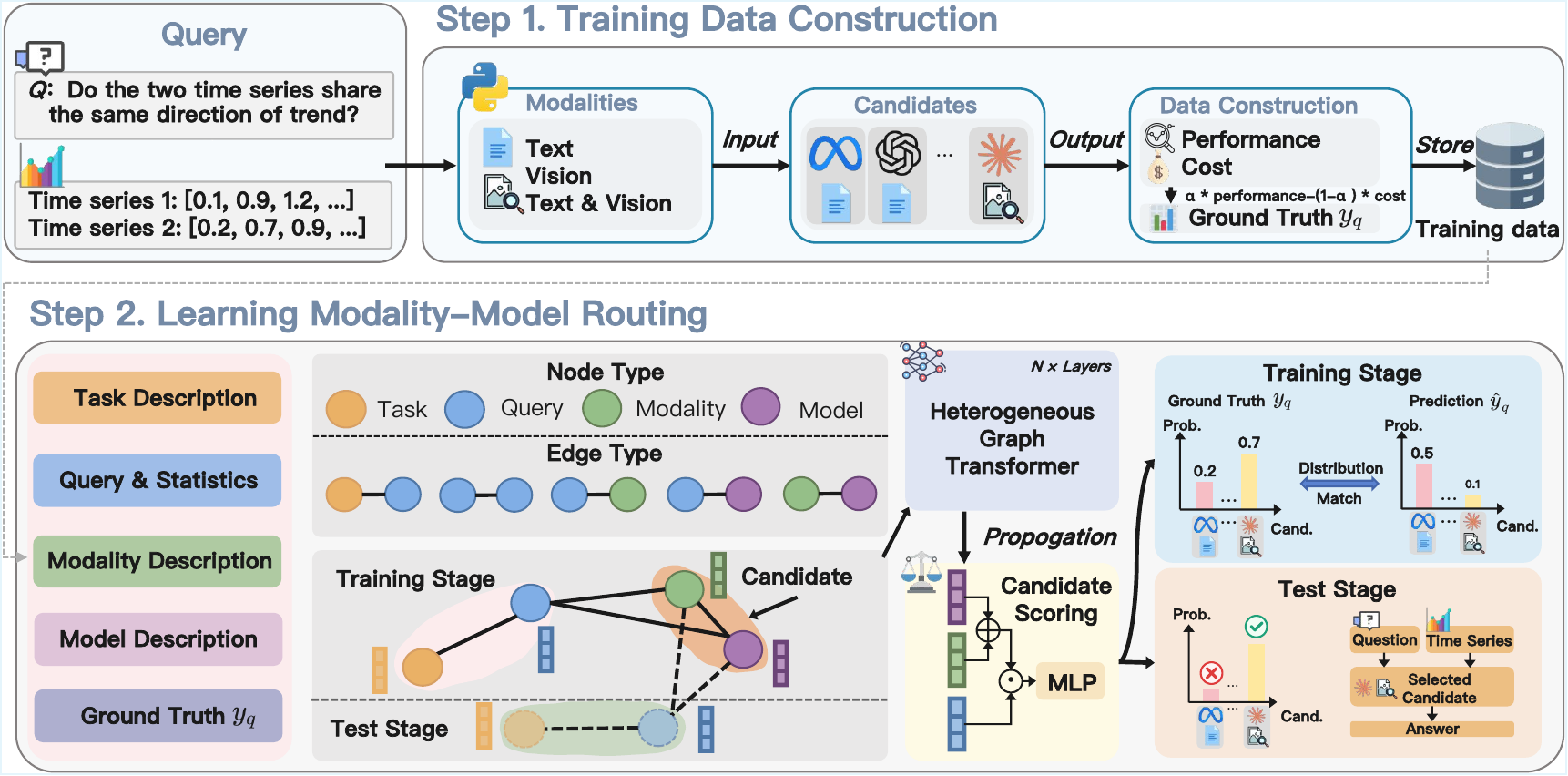}

\caption{Overview of \ours. Each query is paired with all modality--model combinations to collect performance and cost, forming the training data. Then, \ours builds a heterogeneous graph from task, query, modality, and model descriptions, learns node representations, and scores candidates to route queries to the optimal modality--model pair.}

\vspace{-10pt}
\label{fig: overview}
\end{figure*}

\textbf{Overview.} As shown in Figure~\ref{fig: overview}, \ours first transforms the interaction data among tasks, queries, modalities, and models into a graph, and models them as task nodes, query nodes, modality nodes, and model nodes, while the relationships derived from the interaction data are represented by edges. We apply the GNN to embed the node features and use them for routing.
We exploit its compositional structure by constructing a heterogeneous graph $\mathcal{G} = (\mathcal{V}, \mathcal{E})$.

\textbf{Node set.}
The node set $\mathcal{V} = \mathcal{T} \cup \mathcal{Q} \cup \mathcal{M} \cup \mathcal{L}$ comprises 4 disjoint types.
All node features are initialized by encoding natural-language descriptions using a pretrained text-embedding model $\phi$, such as Qwen3-embedding~\citep{zhang2025qwen3}. 
Specifically, (i) \textbf{task nodes} embed a category-level description of scope and requirements, (ii)
\textbf{query nodes} embed the concatenation of the question text $u_i$ and a textual statistical summary of its time series (e.g., channel count, trend direction), allowing the router to consider modality preferences from data characteristics without accessing the raw series (iii)
\textbf{modality nodes} embed a description of modality properties, 
and (iv) \textbf{model nodes} embed a capability-aware profile summarizing architecture, parameter scale, and cost. See all profiles in Appendix~\ref{app: description}.

\paragraph{Edge set.}
The heterogeneous edge set $\mathcal{E} = \mathcal{E}_{\text{TQ}} \cup \mathcal{E}_{\text{QM}} \cup \mathcal{E}_{\text{QL}} \cup \mathcal{E}_{\text{MM}} \cup \mathcal{E}_{\text{QQ}}$ consists of five relation types, each encoding a distinct relationship:
(i)~\textbf{task--query} edges ($\mathcal{E}_{\text{TQ}}$) link each query $q_i$ to its task category $\tau_q(q_i)$ so that queries belonging to the same task share task-level context;
(ii)~\textbf{query--modality} edges ($\mathcal{E}_{\text{QM}}$) link each query $q_i$ to every modality $m_j$;
(iii)~\textbf{modality--model} edges ($\mathcal{E}_{\text{MM}}$) connect each modality $m$ to every compatible model $\ell$, reflecting the valid candidates in $\mathcal{C}$;
(iv)~\textbf{query--model} edges ($\mathcal{E}_{\text{QL}}$) link each query $q_i$ to every model $\ell_k$.
The fifth type captures query similarity:
(v)~\textbf{query--query} edges ($\mathcal{E}_{\text{QQ}}$) connect each query to its $k$-nearest neighbors based on query embeddings' cosine similarity, enabling similar queries to share information during message passing.

\textbf{Routing via Heterogeneous GNN.}
\label{ssec:prediction}
After constructing the interaction histories into a heterogeneous graph, we perform representation learning over $\mathcal{G}$ using a stack of Heterogeneous Graph Transformer (HGT) layers~\citep{hu2020heterogeneous} as our backbone due to its outstanding ability to maintain dedicated representations for different types of nodes.  

To enable cross-type message passing, we encode each node's natural language description with $\phi$ to a embedding space, then apply a linear projection to map them into a shared latent space as the initial features: $\mathbf{h}^{(0)}_v = \mathbf{W} \phi(\mathbf{x}_v) + \mathbf{b}$. Starting from the features $\mathbf{h}^{(0)}_v$, HGT updates
node embeddings by attending to type-specific neighbors, thereby capturing structured interaction
patterns among different types of nodes. Formally, at each layer $l$, the embedding of a node $v$ is
updated by aggregating messages from its neighbors based on relation-aware multi-head attention. We apply residual connections to preserve the initial features and stabilize optimization:
\begin{equation}
    \mathbf{h}^{(\ell+1)}_v = \mathbf{h}^{(\ell)}_v + \text{HGTConv}\!\left(\left\{\mathbf{h}^{(\ell)}_u : (u, r, v) \in \mathcal{E}\right\}\right),
    \label{eq:hgt}
\end{equation}
where $\mathbf{h}^{(l)}_v$ denotes the embedding of node $v$ at layer $l$. After $L$ layers’ propagation, we obtain the final node representations $\mathbf{h}^{(L)}_v$.

To score a candidate $(m, \ell)$ for query $q$, we first combine the modality and model representations additively. A candidate's capability is jointly determined by how the time series is presented and which model processes it. We then capture the interaction between the query and the candidate through element-wise multiplication, followed by a two-layer MLP:
\begin{equation}
    \label{eq:score}
    s(q, m, \ell) = \textsc{MLP}\!\left(\mathbf{h}^{(L)}_q \odot \bigl(\mathbf{h}^{(L)}_m + \mathbf{h}^{(L)}_\ell\bigr)\right),
\end{equation}

Rather than hard labels that only identify the single best candidate, we train with \emph{soft targets} that capture the full ranking over candidates. For each training query $q$, we compute a target distribution by applying softmax to the ground-truth effectiveness scores:
\begin{equation}
    \label{eq:soft_label}
    y_{q,c} = \frac{\exp\bigl(e(q,c)\bigr)}{\sum_{c' \in \mathcal{C}}\exp\bigl(e(q,c')\bigr)},
\end{equation}
Similarly, the predicted scores are converted into a distribution over candidates:
\begin{equation}
    \label{eq:pred_dist}
    \hat{y}_{q,c} = \frac{\exp\bigl(s(q, m, \ell)\bigr)}{\sum_{c' = (m',\ell') \in \mathcal{C}}\exp\bigl(s(q, m', \ell')\bigr)},
\end{equation}
The model is trained by minimizing the KL divergence between the target distribution $y_q$ and the predicted distribution $\hat{y}_q$:
\begin{equation}
    \label{eq:loss}
    \mathcal{L} = \mathcal{D}_{\mathrm{KL}}\!\left(y_q \;\|\; \hat{y}_q\right),
\end{equation}
This objective encourages the router to match the full ranking of candidates rather than merely identifying the top one, providing rich supervision when multiple candidates perform comparably. At test time, we select the candidate with the highest predicted score:
\begin{equation}
    c^* = \arg\max_{(m,\ell) \in \mathcal{C}} \; s(q, m, \ell),
\end{equation}
During training, the graph contains only training queries. At test time, each test query is inserted into the graph with its k-nearest-neighbor edges to existing nodes, and a single forward pass produces the routing score.

\textbf{\ours for new tasks and models.} To incorporate a new task or model, we simply insert the corresponding node into the graph with its embedded feature and connect it following the same edge construction rules. For a new task, \ours then scores all candidates of the existing modality-model against queries in that task. For a new model, \ours leverages messages from neighboring nodes to estimate its per-query effectiveness and incorporate it into routing decisions.
\section{Experiments}
\label{sec: exp}
\subsection{Experimental Setups}
\textbf{Datasets.} We evaluate on \textit{TSRBench}~\citep{yu2026tsrbench}, a comprehensive testbed covering 4 time series reasoning tasks including \textit{Perception}, \textit{Reasoning}, \textit{Prediction}, and \textit{Decision-Making}, with 15 different subtasks, whose statistics are summarized in Table~\ref{tab: stats}. We use two additional tasks as unseen test sets to assess the generalization of \ours, including \textit{time series imputation} from \textit{TSQA} dataset~\citep{kong2025time}, which predicts the missing value in the time series, and \textit{Correlation Prediction} from \textit{MTBench}~\citep{chen2025mtbench}, which requires predicting the correlation between the financial news sentiment and stock price time series.

\noindent\textbf{LLMs and VLMs.} For LLMs, we use Qwen3-8B, Qwen3-32B~\citep{yang2025qwen3}, LlaMa-3.3-70B-Turbo~\citep{grattafiori2024llama}, and Qwen3.5-397B-A17B~\citep{qwen3.5}. For VLMs, we use Qwen3-VL-8B, Qwen3-VL-32B~\citep{bai2025qwen3}, GLM-4.5V~\citep{hong2025glm}, and Kimi-K2.5~\citep{team2026kimi}.

\begin{wraptable}{r}{0.38\textwidth}
\vspace{-15pt}
\centering
\small
\begin{tabular}{lcc}
\toprule
\textbf{Task} & \textbf{\#Subtask} & \textbf{\#Cases} \\
\midrule
Perception & 4 & 700\\
Reasoning & 7 & 1710\\
Prediction & 2 & 1080\\
Decision & 2 & 635\\
\bottomrule
\end{tabular}
\vspace{-5pt}
\caption{Overview of Dataset.}
\label{tab: stats}
\vspace{-15pt}
\end{wraptable}
\noindent\textbf{Data Preprocessing;}
Data are divided into train, validation, and test sets using a 7:1:2 ratio per task. Based on the training split, we construct the training data for \ours by recording per-query correctness and cost (computed from token counts and per-token pricing; see Appendix~\ref{app: description}). Qwen3.5-397B-A17B and Kimi-K2.5 are selected as new LLM and VLM to assess generalizability.

\textbf{Baselines.}
We compare against two groups of baselines. 
(i) \textbf{Rule-based:} \textbf{Largest LLM/VLM} always selects the largest available model. 
(ii) \textbf{Learning-based:} 
\textbf{Hybrid LLM}~\citep{ding2024hybrid} trains an MLP on frozen query embeddings; 
\textbf{RouterDC}~\citep{chen2024routerdc} ranks candidates by cosine similarity between query and model embeddings; 
\textbf{CausalLMRouter}~\citep{ong2024routellm} fine-tunes a decoder to generate the optimal candidate name; 
\textbf{EloRouter}~\citep{ong2024routellm} routes all queries to the highest Elo-rated model; 
\textbf{MFRouter}~\citep{ong2024routellm} uses matrix factorization to score query--model compatibility; 
and \textbf{KNNRouter} selects the best-performing candidate among $k$-nearest training queries.
We adapt all baselines to our joint modality-model routing setting by treating each (modality, model) pair as an independent routing candidate.

\textbf{Evaluation Metrics.} We evaluate each routing method along two axes.
\textit{Accuracy}: following~\citep{yu2025ts, xie2024chatts}, we use accuracy for time series reasoning and correlation prediction tasks; for time series imputation, we follow~\citep{du2024tsi, yildiz2022multivariate} to report Mean Squared Error (MSE) and Mean Absolute Error (MAE).
\textit{Cost}: we report the total API cost in USD consumed on each task's test set, reflecting the expense of deploying each routing method.

\textbf{Implementation Details.}
In the training stage, we search the hyperparameters on the validation set and use a two-layer GNN with an embedding dimension of 64 and $k=60$ for nearest neighbor edges, and train the GNN using Adam optimizer ~\citep{kingma2014adam} with the learning rate as 1e-3. We implement our method using PyTorch Geometric~\citep{fey2019fast} and conduct all experiments on a single NVIDIA RTX A6000 GPU. For LLMs and VLMs, we rely on API calling to obtain responses. 

\subsection{Main Results}
Table~\ref{tab: main results} compares \ours against ten baselines across 4 task categories. \ours achieves 51.33\% overall accuracy, outperforming all routers by 16--46\% relative at competitive cost. Two patterns stand out: (1) always selecting the largest LLM or VLM yields only mediocre accuracy, confirming that scale alone is insufficient; (2) routers with pretrained language model backbones (Hybrid LLM, RouterDC, CausalLM) outperform those without (EloRouter, MFRouter, KNNRouter), yet still lag \ours by a substantial margin. This suggests that query-level semantics benefit routing but are insufficient on their own, and the structured relational signals captured by our heterogeneous graph are critical for joint modality-model selection.

\begin{table*}[t]
\centering

\renewcommand{\arraystretch}{1.1}
\resizebox{\textwidth}{!}{
\begin{tabular}{lcccccccccc}
\toprule
\multirow{2}{*}{\textbf{Scenario}} & 
\multicolumn{2}{c}{\textbf{Perception}} & 
\multicolumn{2}{c}{\textbf{Reasoning}} & 
\multicolumn{2}{c}{\textbf{Prediction}} & 
\multicolumn{2}{c}{\textbf{Decision-Making}} &
\multicolumn{2}{c}{\textbf{Overall}} \\
\cmidrule(lr){2-3} 
\cmidrule(lr){4-5} 
\cmidrule(lr){6-7} 
\cmidrule(lr){8-9}
\cmidrule(lr){10-11}
& \textbf{Accuracy} 
& \textbf{Cost} 
& \textbf{Accuracy} 
& \textbf{Cost} 
& \textbf{Accuracy} 
& \textbf{Cost}
& \textbf{Accuracy} 
& \textbf{Cost}
& \textbf{Accuracy}
& \textbf{Cost} \\
\midrule
\rowcolor{barGray}\multicolumn{11}{c}{\textit{Rule-based Routing Methods}} \\\midrule
LLM (largest) & 48.92 & 0.02 & 33.15 & 0.34 & 31.56 & 0.16 & 34.91 & 0.21 & 35.59 & 0.74 \\
VLM (largest) & \underline{64.03} & 0.03 & 38.76 & 0.29 & 24.44 & 0.22 & 38.68 & 0.05 & 39.10 & 0.59 \\
\midrule
\rowcolor{barGray}\multicolumn{11}{c}{\textit{Learning-based Routing Methods}} \\\midrule
EloRouter & 50.36 & 0.03 & 30.90 & 0.38 & 32.00 & 0.16 & 35.85 & 0.16 & 35.11 & 0.75 \\
MFRouter & 50.36 & 0.03 & 32.58 & 0.38 & 32.00 & 0.16 & 39.62 & 0.17 & 36.32 & 0.77 \\
KNNRouter & 56.12 & 0.03 & 35.11 & 0.34 & 32.44 & 0.19 & \underline{40.57} & 0.17 & 38.62 & 0.79 \\
GraphRouter & 62.58 & 0.03 & 42.42 & 0.38 & 30.22 & 0.20 & 39.62 & 0.16 & 42.13 & 0.77\\
Hybrid LLM & 58.99 & 0.03 & \underline{48.03} & 0.43 & 30.22 & 0.17 & \underline{40.57} & 0.13 & \underline{44.07} & 0.76 \\
RouterDC & 61.87 & 0.03 & 34.27 & 0.41 & \underline{40.00} & 0.23 & 39.62 & 0.17 & 41.16 & 0.83 \\
CausalLM & 55.40 & 0.03 & 39.04 & 0.39 & 38.67 & 0.22 & 39.62 & 0.14 & 41.77 & 0.78 \\
Router-R1 & 58.99 & 0.03 & 38.48 & 0.40 & 33.33 & 0.17 & 35.85 & 0.15 & 40.55 & 0.75 \\
\midrule 
\rowcolor{lightblue!20}
\ours & \textbf{67.63} & 0.03 & \textbf{52.81} & 0.45 & \textbf{43.11} & 0.21 & \textbf{42.45} & 0.04 & \textbf{51.33} & 0.73 \\
\bottomrule
\end{tabular}
}
\vspace{-5pt}
\caption{Performance and cost comparison of various baselines on 4 time series reasoning tasks. Bold and underline denote the best and second-best results.}
\label{tab: main results}
\end{table*}

\subsection{Generalization Ability of \ours}
\begin{figure}[t]
    \centering
    \begin{minipage}{0.55\textwidth}
        \centering
        \includegraphics[width=\linewidth]{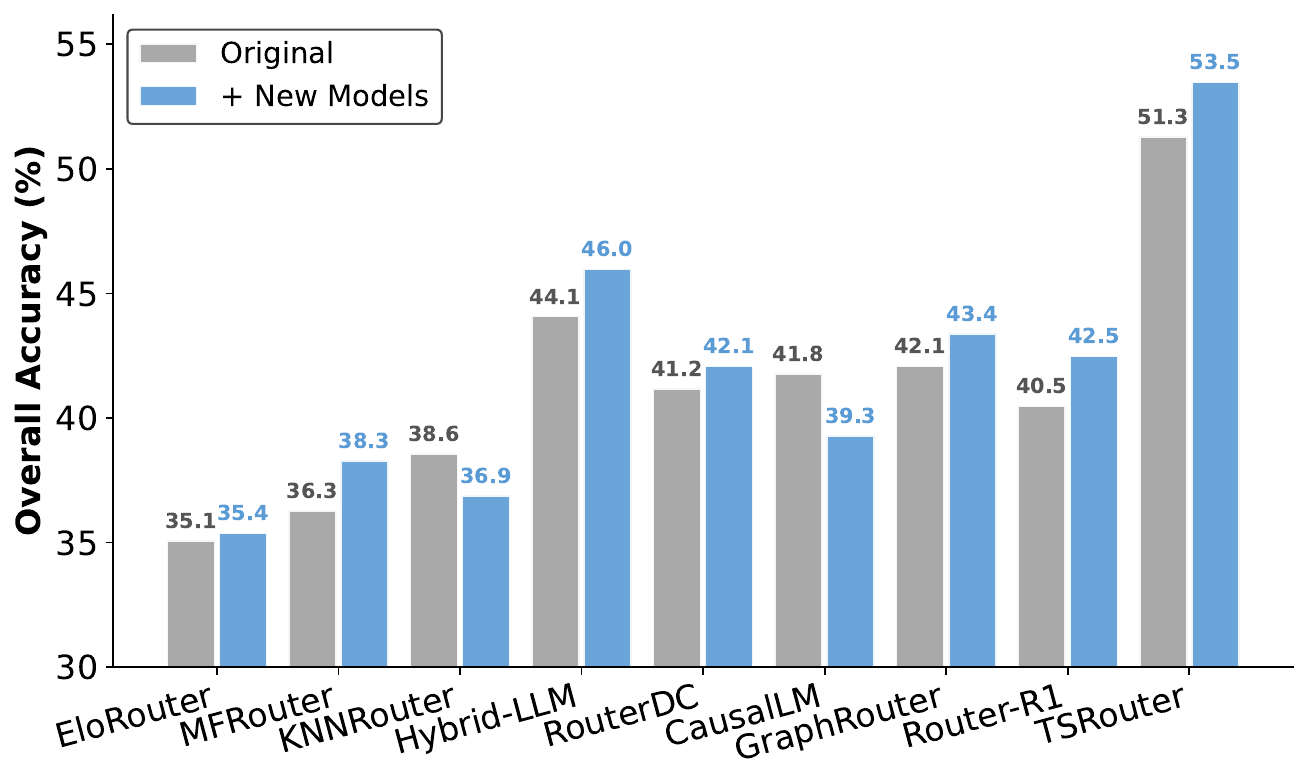}
        \caption{Results on generalization to new models.}
        \label{fig:new_models}
    \end{minipage}
    \hfill
    \begin{minipage}{0.44\textwidth}
        \centering
        \small
        \resizebox{\textwidth}{!}{
        \begin{tabular}{lccc}
        \toprule
        \multirow{2.5}{*}{\textbf{Scenario}} & 
        \textbf{Correlation Prediction} & 
        \multicolumn{2}{c}{\textbf{Imputation}} \\
        \cmidrule(lr){2-2} 
        \cmidrule(lr){3-4} 
        & \textbf{Accuracy} 
        & \textbf{MSE} 
        & \textbf{MAE}  \\
        \midrule
        LLM (largest) & 22.24 & 0.67 & 0.46 \\
        VLM (largest)   & 29.20 & 0.89 & 0.46 \\
        \midrule
        EloRouter  & 25.42 & 1.07 & 0.65  \\
        MFRouter   & 25.22 & 1.07 & 0.65 \\
        KNNRouter  & 24.93 & 0.99 & 0.59\\ 
        GraphRouter & 27.51 & 0.61 & 0.49\\
        Hybrid LLM & 29.59 & 0.60 & 0.46 \\
        RouterDC   & 28.60 & 0.74 & 0.47 \\
        CausalLM   & 27.81 & 0.89 & 0.47 \\
        Router-R1  & 29.59 & 0.70 & 0.48 \\
        \midrule 
        \rowcolor{lightblue!20}
        \ours & \textbf{31.38} & \textbf{0.56} & \textbf{0.43} \\
        \bottomrule
        \end{tabular}
        }
        \captionof{table}{Results on generalization to new time series tasks.}
        \label{tab: new tasks}
    \end{minipage}
    \vspace{-10pt}
\end{figure}
\textbf{Generalization to New Models.} A critical requirement for any practical routing framework is the ability to incorporate newly released models efficiently. We expand the candidate pool with two models (one LLM, one VLM) \textit{unseen} during training and compare all routers under both the original and augmented pools.
As illustrated in Figure~\ref{fig:new_models}, \ours achieves the largest gain, increasing from 51.3\% to 53.5\%. In contrast, KNNRouter and CausalLM exhibit performance degradation upon incorporating new models, indicating that these methods fail to leverage the capabilities of unseen models and instead disrupt their existing routing decisions. These results confirm that \ours can effectively integrate new models without retraining.

\textbf{Generalization to New Tasks.}
Beyond the 4 time series reasoning tasks, we evaluate \ours on two held-out time series tasks: \textit{correlation prediction}, which requires jointly analyzing news sentiment and stock price time series to assess their correlation, and \textit{time series imputation}, which requires predicting missing numerical values in a sequence.
As shown in Table~\ref{tab: new tasks}, \ours consistently outperforms all baselines on both tasks. Notably, the two tasks exhibit opposite modality preferences: the largest VLM leads on correlation prediction, where visual patterns aid alignment, while the largest LLM leads on imputation, where numerical precision is essential. Most baseline routers perform poorly on both tasks, indicating weak generalization to unseen task types. These results confirm that \ours effectively identifies the characteristics of novel tasks and routes queries accordingly.

\subsection{Modality Routing Analysis}
To understand how \ours allocates queries for different time series modalities, we analyze the proportion of instances routed to each modality per task category. 
\begin{wrapfigure}{r}{0.60\textwidth}
    \vspace{-10pt}
	\centering
	\includegraphics[width=\linewidth]{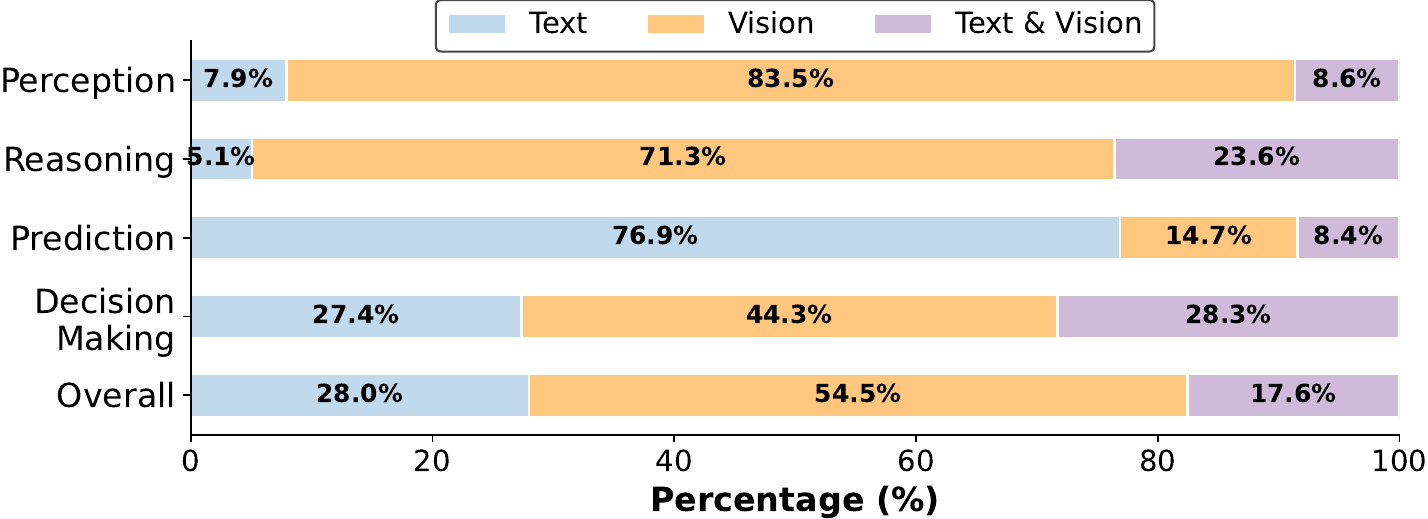}
	\caption{Modality routing distribution across tasks of \ours.}
    \vspace{-5pt}
	\label{fig:modality_routing}
\end{wrapfigure}
As shown in Figure~\ref{fig:modality_routing}, \ours learns routing strategies that align closely with the modality distribution identified in \S\ref{sec: motivation} and results in Table~\ref{tab: main results}.
Perception and reasoning queries are predominantly routed to the vision modality. 
Prediction queries, by contrast, are overwhelmingly routed to textual modality, reflecting the need for precise numerical extrapolation that textual modality better preserves. 
Decision-making exhibits the most balanced distribution across all three modality types, reflecting its need for both quantitative precision and global pattern understanding. These results demonstrate that \ours effectively captures the modality demands of different queries and tasks.

\begin{figure*}[t]   
\centering

\includegraphics[width=1\columnwidth]{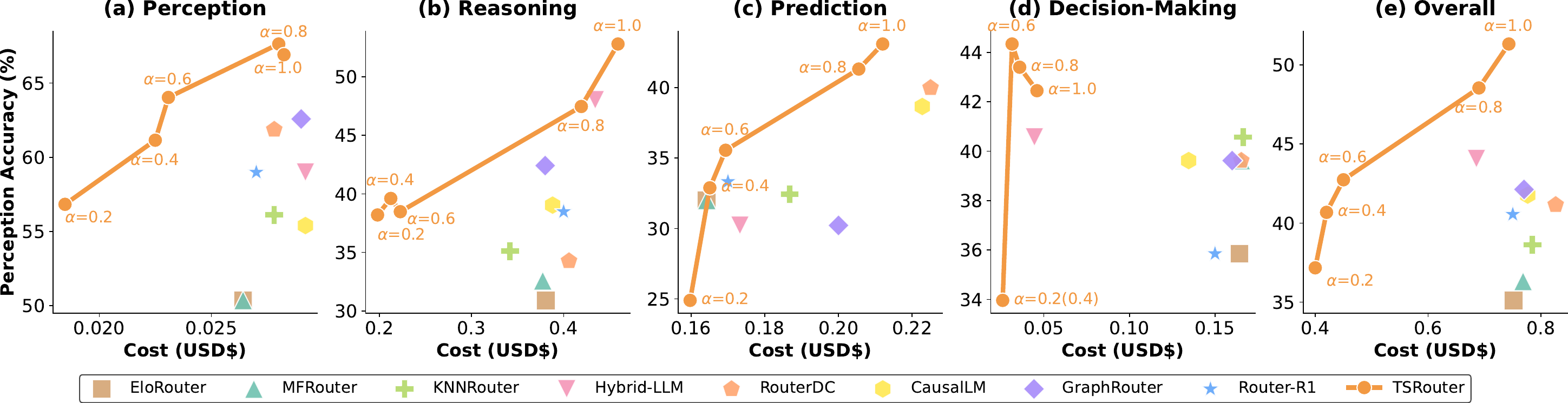}

\caption{Accuracy--cost trade-off under varying cost preference $\alpha$. Each subplot corresponds to a task and overall accuracy. \ours traces a clear Pareto front, consistently dominating all baselines across the full range of cost budgets.}

\label{fig:cost_tradeoff}
\vspace{-10pt}
\end{figure*}

\subsection{Routing under Cost Constraints}
In real-world deployment, a practical routing system must balance performance against inference cost. To investigate the impact of costs in routing, we train \ours at varying cost coefficients $\alpha$ in Equation~\ref{eq:effectiveness} and compare against various baselines. The Pareto-style plots in Figure~\ref{fig:cost_tradeoff} illustrate a clear trade-off between the answer accuracy and computational cost. When $\alpha=1.0$, \ours achieves the highest accuracy across all tasks, reflecting its performance-oriented routing preference. As $\alpha$ decreases, \ours gradually shifts toward more cost-efficient decisions, substantially reducing the cost but with a moderate decline in performance. Notably, at a cost budget comparable to the best-performing baseline (Hybrid LLM), \ours still significantly outperforms all baselines in accuracy, indicating that \ours achieves both high performance and cost efficiency, providing practitioners with flexible control over the accuracy-cost operating point.

\subsection{Ablation Studies}
\begin{figure*}[t]   
	\centering
	\includegraphics[width=1.0\textwidth]{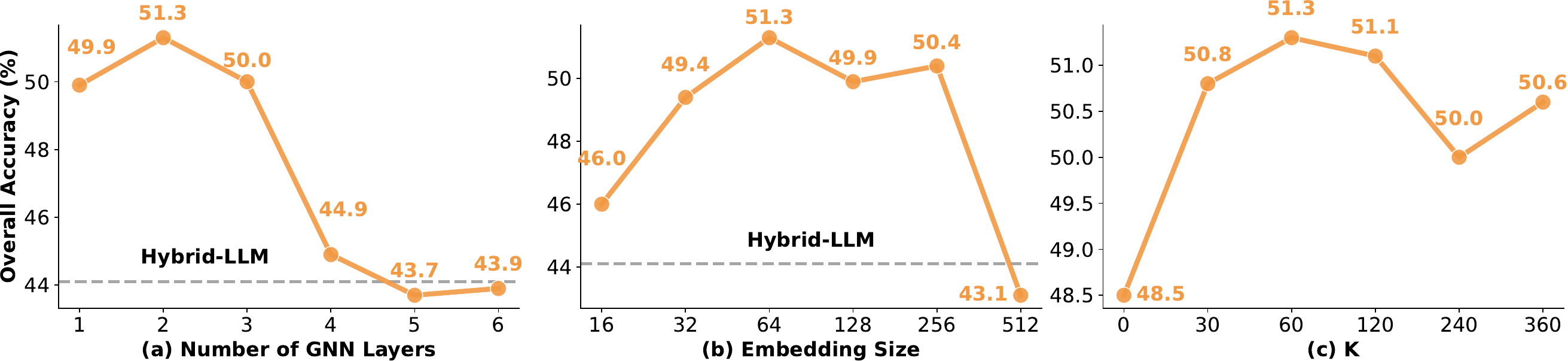}
	\caption{Hyperparameter sensitivity analysis. (a) Number of HGT layers; (b) Embedding dimension; (c) Number of nearest neighbors $k$. The dashed line indicates the best-performing baseline.}
	\label{fig: hyperparams}
	\vspace{-15pt}
\end{figure*}

\textbf{Effect of the Number of GNN Layers.} The number of GNN layers has a significant impact on the expressiveness of the GNN. A shallow GNN struggles to learn deep contextual information, whereas an overly deep GNN results in performance degradation issues such as oversmoothing~\citep{kipf2016semi}. We conduct an exploration with the number of GNN layers for \ours, ranging from 1 to 6. As shown in Figure~\ref{fig: hyperparams} (a), the performance initially improves with more layers but then declines significantly, achieving its highest performance when the number of layers is 2.

\begin{wrapfigure}{r}{0.45\textwidth}
    \vspace{-10pt}
	\centering
	\includegraphics[width=\linewidth]{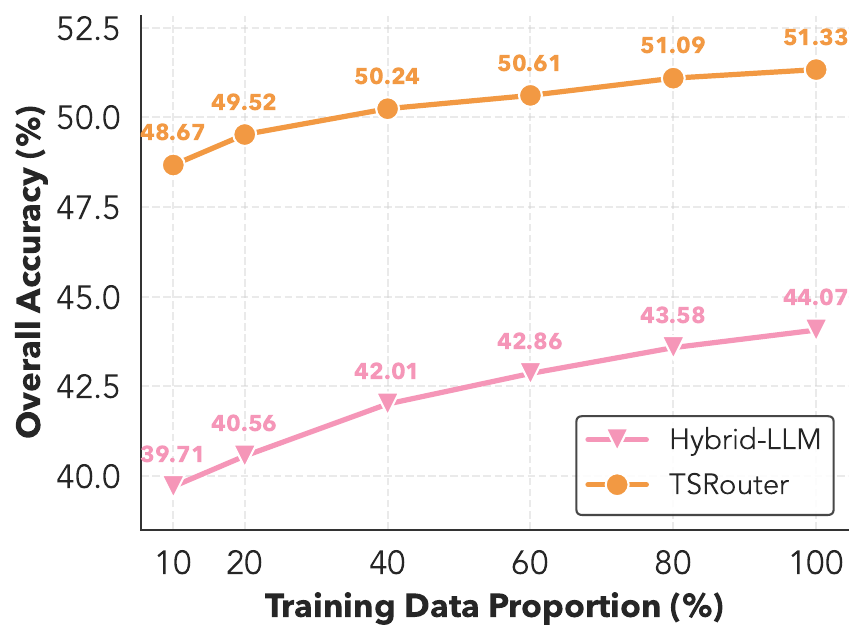}
	\caption{Performance of \ours using different amounts of training data.}
    \vspace{-5pt}
	\label{fig:data scaling}
	\vspace{-10pt}
\end{wrapfigure}

\textbf{Effect of the Amount of Training Data.} We further examine how \ours performs under varying amounts of training data, comparing against the strongest baseline. As shown in Figure~\ref{fig:data scaling}, \ours trained on only 10\% of the data already surpasses the best baseline trained on the full dataset by a large margin, demonstrating the strong data efficiency of our approach.

\textbf{Effect of the Embedding Size of GNN.} The size of a GNN is an important factor that affects both the performance and computational overhead. Figure~\ref{fig: hyperparams} shows that the performance of \ours initially improves as the embedding size increases, reaching its peak at a size of 64, after which it starts to decline.

  \begin{wraptable}{r}{0.69\textwidth}
    \centering
    \small
    \begin{tabular}{lccccc}
    \toprule
    \textbf{Components} & 
    \textbf{Perc.} & 
    \textbf{Reas.} & 
    \textbf{Pred.} & 
    \textbf{Deci.} &
    \textbf{Over.} \\
    \midrule
    \ours (Full) & \textbf{67.63} & \textbf{52.81} & \textbf{43.11} & \textbf{42.45} & \textbf{51.33} \\
    \midrule
    - Hetero Graph & 64.03 & 49.44 & 41.78 & 37.74 & 46.25 \\
   - Query-Query Edges & 65.47 & 49.72 & 39.11 & 42.45 & 48.55 \\
    - MLP Scoring Head  & 64.03 & 44.10 & 40.44 & 41.51 & 46.52 \\
    - Modality-Model Edges & 66.91 & 49.44 & 36.44 & 37.77 & 47.33 \\
    \bottomrule
    \end{tabular}
    \vspace{-5pt}
    \caption{Ablation study of \ours components.}
    \label{tab:ablation}
    \vspace{-10pt}
    \end{wraptable}
\textbf{Effect of Components in \ours.} We conduct a series of ablation studies on \ours along the following aspects: (i) \textbf{Homogeneous Graph}: setting all edge types identical to reduce to a homogeneous graph, evaluating the benefit of heterogeneous message passing.
    (ii) \textbf{Query-Query Connections}: removing the nearest query--query edges ($\mathcal{E}_{\text{QQ}}$) to assess the value of inter-query information sharing; 
    (iii) \textbf{Dot-Product Prediction Head}: replacing the MLP-based prediction head (Eq.~\ref{eq:score}) with a simple dot-product;
    (iv) \textbf{Interactions within Modality-Models}: we merge each modality--model nodes into a single candidate node, eliminating the separate additive composition and the modality--model structural edges.
The results shown in Table~\ref{tab:ablation} highlight the criticality of each element in \ours.


\section{Related Work}
\textbf{LLMs/VLMs for Time Series.} Beyond traditional statistical methods~\citep{rb1990stl, liu2021pyraformer, lim2021temporal, wu2021autoformer, li2023prototype, li2024transformer}, recent work adapts LLMs for time series via prompting~\citep{cao2023tempo, ltsm-bundle, guan2025timeomni} and reprogramming~\citep{jin2023time, pillai2025time2lang}, while a parallel line leverages VLMs by converting time series into visual plots~\citep{chen2024visionts, zhong2025time, Liu2024API, jiang2025timexl, guan2026timeomni}. LLMs and VLMs exhibit complementary strengths~\citep{yu2026tsrbench, kong2025achieving}: raw numbers excel at precise numerical understanding, while visualizations better capture global patterns. Moreover, individual models vary in expertise, solving different subsets of problems even within the same task. We exploit this complementarity through dynamic, query-level routing that jointly selects the optimal modality and model.

\textbf{LLM Router.} LLM routers balance efficiency and accuracy by selecting among multiple LLMs per query, typically via query embeddings or trained classifiers, such as RouterKNN~\citep{shnitzer2023large}, RouterMLP~\citep{shnitzer2023large}, RouterSVM~\citep{hu2024routerbench}, RouterDC~\citep{chen2024routerdc}, RouteLLM~\citep{ong2024routellm}, GraphPlanner~\citep{feng2026graphplanner}, GraphRouter~\citep{feng2024graphrouter, llmrouter2025}, and Hybrid LLM~\citep{ding2024hybrid}. However, these methods focus solely on model selection under a fixed input modality. For instance, GraphRouter formulates routing as edge prediction between query and model nodes, which does not naturally extend to joint modality-model selection. We go beyond by jointly optimizing both dimensions, routing each query to the best modality-model combination.

\textbf{LLM/VLM Reasoning.} Recent advances in LLMs and VLMs have substantially improved their capacity for deliberate problem solving. Prompting approaches elicit intermediate reasoning steps~\citep{wei2022chain, zhou2022least, ho2025arcmemo, yu2025hallurnn, min2026stop}, but their effectiveness is bounded by the model's intrinsic abilities. Subsequent work therefore turns to test-time compute to increase reasoning depth and broaden exploration of the reasoning space, including sampling strategies~\citep{brown2024large, karan2025reasoning, aggarwal2023let} and structured reasoning frameworks~\citep{yao2023tree, besta2024graph, hao2023reasoning, ding2024everything}. More recently, a growing line of research internalizes reasoning capabilities through reinforcement learning~\citep{yu2024flow, yu2026arrowgev, cheng2026revisiting, guo2025deepseek} and on-policy distillation~\citep{zhao2026self, yang2026self, shenfeld2026self, li2026rethinking}, but at the cost of substantial training effort and computational resources. \ours sits at the intersection of LLM and VLM reasoning, which offers a lightweight solution that leverages the strengths of both for time series reasoning with only minimal training cost.

\vspace{-10pt}
\section{Conclusion}
\vspace{-5pt}
We introduce \ours, a dynamic routing framework that optimizes both modality and model selection for time series reasoning. Through a heterogeneous graph that captures interactions among tasks, queries, modalities, and models, \ours outperforms diverse baselines across 4 reasoning tasks in both accuracy and cost-efficiency, while generalizing to unseen tasks and models without retraining.

\bibliography{colm2026_conference}
\bibliographystyle{colm2026_conference}

\appendix
\newpage
\section{Additional Experimental Results}

\label{app:details}
\begin{figure}[t]
    \centering
    \begin{minipage}{0.49\textwidth}
        \centering
        \small
        \resizebox{\textwidth}{!}{
        \begin{tabular}{lccccc}
        \toprule
        \textbf{Variant} & 
        \textbf{Perc.} & 
        \textbf{Reas.} & 
        \textbf{Pred.} & 
        \textbf{Deci.} &
        \textbf{Over.} \\
        \midrule
        HGT & 67.63 & 52.81 & 43.11 & 42.45 & 51.33 \\
        HAN & 71.22 & 51.12 & 43.11 & 42.45 & 51.21 \\
        HeteroGAT & 69.06 & 47.75 & 44.89 & 43.40 & 50.00 \\
        \bottomrule
        \end{tabular}
        }
        \captionof{table}{Results with different GNN backbone architectures for \ours.}
        \label{tab:gnn_ablation}
    \end{minipage}
    \hfill
    \begin{minipage}{0.49\textwidth}
        \centering
        \includegraphics[width=\linewidth]{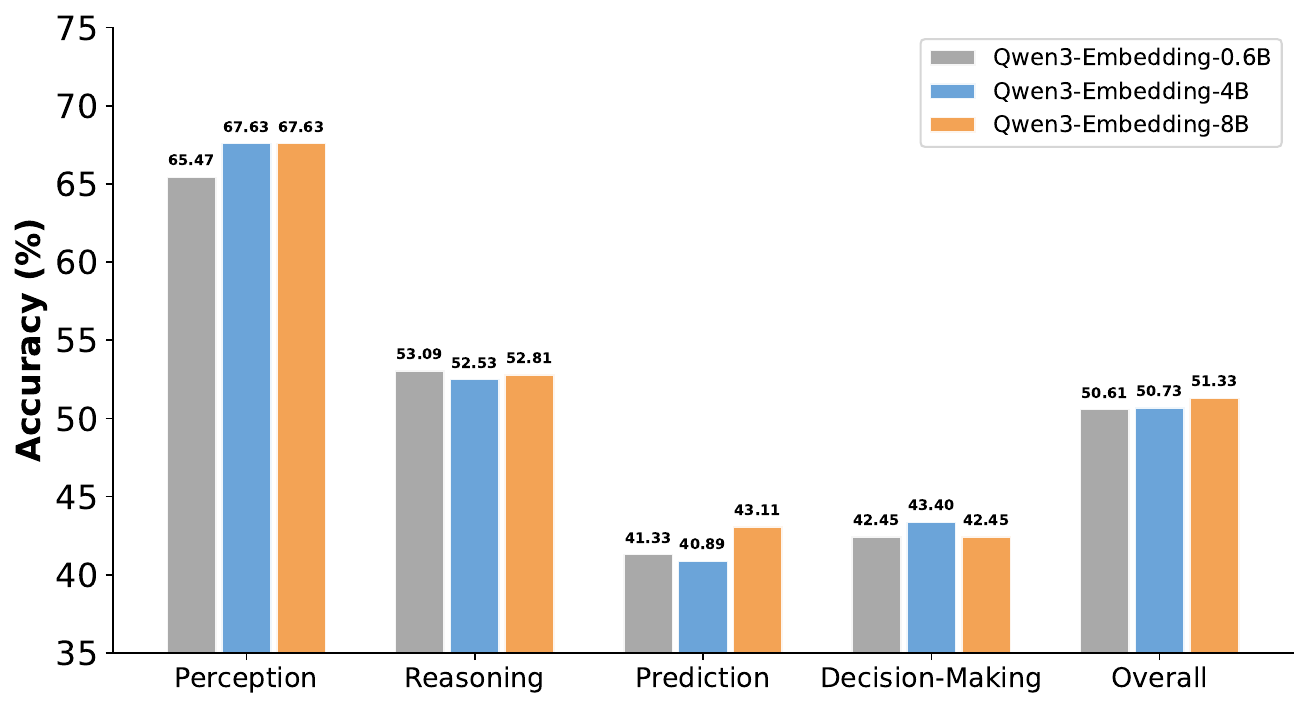}
        \caption{Performance with different embedding models for \ours.}
        \label{fig: model size}
    \end{minipage}
\vspace{-10pt}
\end{figure}

\subsection{Performance under Different GNNs.}
To investigate the performance with different GNN backbones, we replace the HGT with two widely used alternatives, HAN and HeteroGAT, while keeping all other components unchanged. As shown in Table~\ref{tab:gnn_ablation}, all three variants achieve highly competitive overall accuracy, ranging from 50.00\% to 51.33\%, with a gap of merely 1.33 points between the best and worst configurations. 
Notably, no single backbone uniformly dominates across all subtasks: HGT excels at reasoning, HAN achieves the highest perception score, and HeteroGAT leads on both prediction and decision-making. This complementary pattern suggests that the routing performance of \ours is primarily driven by the overall heterogeneous graph formulation, which explicitly model the interaction between contextual information, rather than by the specific message-passing mechanism employed.

\subsection{Performance under Different Embedding Models}
To investigate the robustness of \ours with different embedding model backbones for node feature initialization, we evaluate \ours across three distinct sizes of the Qwen3-Embedding model: 0.6B, 4B, and 8B. The results shown in Figure~\ref{fig: model size} confirm that \ours is both highly effective and robust. 
We observe a positive correlation between overall accuracy and the model size, indicating that the embeddings from larger embedding models are more informative for the following contextual information interactions. Performance remains consistent across all tasks, demonstrating the robustness of the embedding model and representation pipeline. This suggests that the downstream GNN is resilient to moderate variations in input features.

\begin{wraptable}{r}{0.65\textwidth}
    \centering
    \small
    \vspace{-10pt}
    \resizebox{0.65\textwidth}{!}{
    \begin{tabular}{lccccc}
    \toprule
    \textbf{Variant} & 
    \textbf{Perc.} & 
    \textbf{Reas.} & 
    \textbf{Pred.} & 
    \textbf{Deci.} &
    \textbf{Over.} \\
    \midrule
    \makecell[l]{\ours \\ w/o description} & 66.20\scriptsize{$\pm$0.37} & 51.70\scriptsize{$\pm$0.17} & 43.40\scriptsize{$\pm$0.25} & 40.90\scriptsize{$\pm$0.52} & 50.50\scriptsize{$\pm$0.34} \\
    \midrule
    \makecell[l]{\ours \\ w/ description} & 68.35\scriptsize{$\pm$0.72} & 52.34\scriptsize{$\pm$0.59} & 42.81\scriptsize{$\pm$0.26} & 42.77\scriptsize{$\pm$0.55} & 51.21\scriptsize{$\pm$0.21} \\
    \bottomrule
    \end{tabular}
    }
    \caption{Results on variants of node text descriptions for \ours.}
    \label{tab:desc_ablation}
    \vspace{-10pt}
\end{wraptable}

\subsection{Performance when Removing LLM-generated Descriptions}
To assess the robustness of \ours to the quality of node descriptions, we remove the LLM-generated descriptions and initialize node features using only the task, modality, and model names. We report results averaged over 3 independent runs in Table~\ref{tab:desc_ablation}. Even without the descriptions, \ours exhibits only a minor performance drop and still clearly outperforms the best baseline. This is because the descriptions serve merely as an initial semantic prior for node features, whereas the routing signal is primarily produced by structural message passing over the heterogeneous graph. Consequently, \ours remains effective and highly competitive without relying on LLM-generated descriptions.

\subsection{Efficiency Analysis}
In real-world deployment, low routing latency and memory use are essential for responsive systems. We compare \ours with the five strongest learning-based baselines (RouterDC, Hybrid LLM, CausalLM, GraphRouter, and Router-R1) in terms of per-query routing time and memory usage (Table~\ref{tab:efficiency}). \ours achieves the lowest latency at 
\begin{wraptable}{r}{0.7\textwidth}
\vspace{-5pt}
\centering
\begin{tabular}{lcc}
\toprule
\textbf{Method} & \textbf{Inference Time (ms)} & \textbf{Memory Use (MiB)} \\
\midrule
CausalLM & 910.78 & 33400\\
RouterDC & 18.64 & 29358\\
HybridLLM & 27.65 & 29299\\
GraphRouter & 4.02 & 1433\\
Router-R1 & 66.80 & 41177 \\ \midrule
\ours & \textbf{3.21} & \textbf{576}\\
\bottomrule
\end{tabular}
\caption{Per-query routing latency comparison.}
\label{tab:efficiency}
\vspace{-10pt}
\end{wraptable}
3.21\,ms per query, approximately $6\times$ faster than RouterDC, $9\times$ faster than Hybrid LLM, and $284\times$ faster than CausalLM. CausalLM incurs the highest cost as it relies on an LLM itself for routing decisions. More broadly, all three baselines need to process each query independently, limiting parallelism. In contrast, the graph-based formulation of \ours naturally supports batch inference, where multiple queries are incorporated into the graph and the GNN operates over the entire graph simultaneously, amortizing overhead across queries and significantly improving the computational efficiency. The memory advantage stems from \ours's lightweight GNN architecture, whereas most baselines load pretrained language models as their routing backbone.

\section{Additional Setup Details}
\label{app: description}
\textbf{Description Prompts.} Using descriptions generated by LLM and embeddings derived from a pretrained embedding model as initial node features for GNNs enhances their expressiveness and generalization capability. Here, we list descriptions of different tasks, modalities, and models. The profiles contain only the public attributes without the information of the time series benchmark, which ensures there is no data leakage. 
The detailed descriptions are shown in the following figures. 

\textbf{Cost Details.} We report the cost for 1M token input and output. For VLMs, image tokens are billed according to each provider's pricing schedule. We detail the cost information of all the models in Table~\ref{tab: costs}.

\textbf{Time Series Visualization.} Each time series is rendered as a line chart image before being passed to VLMs. Given a query with $N$ channels, we plot each channel as a separate subplot, with a shared x-axis (time/index) and individual y-axes. For univariate series, a single plot is generated; for multivariate series, we stack $N$ subplots vertically, each labeled with its channel name. We apply grid lines and bold axis labels to preserve visual clarity. All charts have 100 PPI resolution.

\begin{table}[t]
\centering
\resizebox{\textwidth}{!}{%
\begin{tabular}{lcccc}
\toprule
\textbf{Model} & \textbf{Type} & \textbf{Input (\$/1M Token)} & \textbf{Output (\$/1M Token)} & Provider \\
\midrule
Qwen3-8B              & LLM & 0.12 & 0.45 & Alibaba Cloud Int.\\
Qwen3-32B             & LLM & 0.16 & 0.64 & Alibaba Cloud Int.\\
LLaMA-3.3-70B-Turbo   & LLM & 0.25 & 0.75 & Together AI\\
Qwen3-VL-8B           & VLM & 0.12 & 0.45 & Alibaba Cloud Int.\\
Qwen3-VL-32B          & VLM & 0.16 & 0.64 & Alibaba Cloud Int.\\
GLM-4.5V              & VLM & 0.20 & 1.10 & Together AI\\
\midrule
Qwen3.5-397B-A17B     & LLM & 0.65 & 3.60 & Together AI\\
Kimi-K2.5             & VLM & 0.50 & 2.80 & Together AI\\
\bottomrule
\end{tabular}
}
\caption{LLMs and VLMs used in \ours with costs per 1M tokens.}
\label{tab: costs}
\end{table}

\input{prompts/case_study}

\begin{table*}[t]
\centering

\scriptsize
\renewcommand{\arraystretch}{1.2}
\begin{tabularx}{\textwidth}{@{}l >{\raggedright\arraybackslash}X c@{}}
\toprule
\textbf{Model (Modality)} & \textbf{Reasoning Process} & \\
\midrule
\rowcolor{red!5}
\textbf{Qwen3-8B} (Text) & \textit{Ans: (B)} --- The time series data shows an initial steady increase (linear trend), followed by a more rapid growth (exponential trend), and eventually a leveling off or slower growth (logarithmic trend). The data starts with a gradual rise, then accelerates, and later appears to stabilize\ldots & \textcolor{red}{\textbf{\ding{55}}} \\
\rowcolor{red!5}
\textbf{Qwen3-32B} (Text) & \textit{Ans: (B)} --- Observing the given data, it first increases at a steady rate (linear trend), then starts growing faster (exponential trend), and finally the growth slows and stabilizes (log trend). & \textcolor{red}{\textbf{\ding{55}}} \\
\rowcolor{red!5}
\textbf{LlaMa-3.3-70B} (Text) & \textit{Ans: (B)} --- The time series starts with a steady increase (linear), followed by a rapid growth (exponential), and then a slowing down with minor fluctuations (logarithmic). & \textcolor{red}{\textbf{\ding{55}}} \\
\midrule
\rowcolor{green!5}
\textbf{Qwen3-VL-8B} (Vision) & \textit{Ans: (D)} --- The plot shows three distinct phases. Initially (time 0 to $\sim$40), the series exhibits rapid, accelerating growth --- characteristic of an exponential trend. After time 40, the growth rate slows significantly\ldots & \textcolor{forestgreen}{\textbf{\ding{51}}} \\
\rowcolor{green!5}
\textbf{Qwen3-VL-32B} (Vision) & \textit{Ans: (D)} --- The plot shows three distinct phases: 1.\ Initial phase (0--40): rapid increase with accelerating rate --- exponential growth. 2.\ Middle phase (40--80): growth becomes constant --- linear trend. 3.\ Final phase: curve flattens --- logarithmic trend. & \textcolor{forestgreen}{\textbf{\ding{51}}} \\
\rowcolor{green!5}
\textbf{GLM-4.5V} (Vision) & \textit{Ans: (D)} --- Three distinct segments: 1) Initial steep upward curve (exponential), 2) consistent upward slope (linear), 3) gradually decreasing slope (logarithmic). & \textcolor{forestgreen}{\textbf{\ding{51}}} \\
\midrule
\rowcolor{yellow!8}
\textbf{Qwen3-VL-8B} (Text+Vision) & \textit{Ans: (D)} --- The data shows a clear progression through multiple trend phases. Initially (to $\sim$40), rapid accelerating growth characteristic of an exponential trend. After index 40, growth slows down significantly\ldots & \textcolor{forestgreen}{\textbf{\ding{51}}} \\
\rowcolor{yellow!8}
\textbf{Qwen3-VL-32B} (Text+Vision) & \textit{Ans: (B)} --- The data starts with a slow increase, then accelerates rapidly around index 40--50, and finally stabilizes. The initial phase shows exponential-like growth, but \underline{the text data biases toward interpreting it as linear}\ldots & \textcolor{red}{\textbf{\ding{55}}} \\
\rowcolor{yellow!8}
\textbf{GLM-4.5V} (Text+Vision) & \textit{Ans: (A)} --- Analyzing the data and visualization, I identify three phases. The initial phase (0--30) shows a linear growth pattern with a steady upward slope. The middle phase (30--80) exhibits exponential growth. The final phase (80--130) displays logarithmic growth\ldots & \textcolor{red}{\textbf{\ding{55}}} \\
\bottomrule
\end{tabularx}
\caption{Reasoning Trace of each modality-model for the case in Figure~\ref{fig:case-study}}
\label{tab:case_study}
\end{table*}

\section{Time Series Statistical Summary}
\label{app:ts_stats}

Each query node embeds the concatenation of the question text $u_i$ and a textual statistical summary of its time series $\mathbf{x}_i$. This summary is appended to the question text before encoding with the pretrained embedding model $\phi$, allowing the router to infer modality preferences from data characteristics.

For each time series $\mathbf{x}_i \in \mathbb{R}^{T_i}$, we compute the following statistics:

\begin{itemize}[leftmargin=*,itemsep=2pt]
    \item \textbf{Mean}: $\bar{x}_i = \frac{1}{T_i}\sum_{t=1}^{T_i} x_{i,t}$.
    \item \textbf{Standard deviation}: $\sigma_i = \sqrt{\frac{1}{T_i}\sum_{t=1}^{T_i}(x_{i,t} - \bar{x}_i)^2}$.
    \item \textbf{Minimum and maximum}: $\min(\mathbf{x}_i)$ and $\max(\mathbf{x}_i)$.
    \item \textbf{Trend direction}: determined by the sign of the slope $\hat{\beta}_1$ from a least-squares linear fit $x_{i,t} \approx \beta_0 + \beta_1 t$, reported as \emph{upward} ($\hat{\beta}_1 > 0$) or \emph{downward} ($\hat{\beta}_1 \leq 0$).
\end{itemize}

\noindent These statistics are rendered into a natural language sentence and appended to the query text before embedding. 
\section{A Case on Reasoning with Different Modalities and Models}
\label{sec: appendix motivation}

We provide a sample question in Figure~\ref{fig:case-study}, and report the reasoning trace of different candidates in Table~\ref{tab:case_study}. All three LLMs operating on textual time series arrive at the wrong answer, misidentifying the initial exponential phase as linear, likely because the early absolute increments appear small in raw numerical form. In contrast, all three VLMs correctly identify the trend ordering from the visual plot, where the curvature of each regime is directly observable. Interestingly, when both modalities are provided simultaneously, the textual numerical values partially interfere with visual reasoning. For example, Qwen3-VL-32B and GLM-4.5V are both correct under vision-only input, but flip to incorrect answers given the additional textual time series. This case highlights clear differences in reasoning across modalities and models. All text-only LLMs arrive at the wrong answer, while all vision-only VLMs succeed. Furthermore, the smaller Qwen3-VL-8B answers correctly under mixed input while the larger Qwen3-VL-32B does not, confirming that larger models are not always optimal. These observations motivate \ours to jointly select the best modality and model per query.
 
\input{prompts/tasks/perception}

\input{prompts/tasks/reasoning}

\input{prompts/tasks/prediction}

\input{prompts/tasks/decision_making}

\input{prompts/modalities/text}

\input{prompts/modalities/visual}

\input{prompts/modalities/mix}

\input{prompts/models/qwen3-8b}

\input{prompts/models/qwen3-32b}

\input{prompts/models/llama-3.3-70b-instruct-turbo}

\input{prompts/models/qwen3-vl-8b-instruct}

\input{prompts/models/qwen3-vl-32b-instruct}

\input{prompts/models/glm-4.5v}

\input{prompts/models/qwen3.5-397b-a17b}

\input{prompts/models/kimi-k2.5}

\end{document}

%% file: prompts/case_study.tex
\begin{figure}[t]
    \begin{ReasoningBox}[title=Case Study, breakable=false]
        \textbf{Question:} The given time series has multiple trends that follow each other. What is the correct ordering of the trend components?\\
        \begin{wrapfigure}{r}{0.55\textwidth}
            \centering
            \vspace{-15pt} 
            \includegraphics[width=\linewidth]{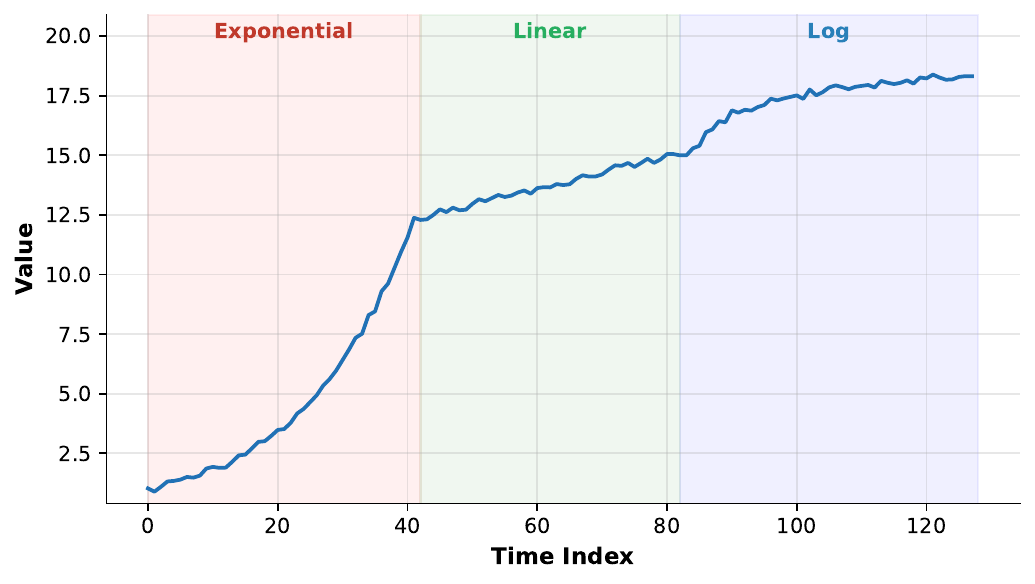}
        \end{wrapfigure}
        
        \textbf{Answer Choices:} \\ 
        (A) Log \\ 
        (B) Linear$\rightarrow$Exponential$\rightarrow$Log \\ 
        (C) Linear→Exponential \\ 
        (D) Exponential→Linear→Log \\ 
        \textbf{Correct Answer:} \textcolor{forestgreen}{(D)} \\
        \vspace{4em}
    \end{ReasoningBox}
    \caption{Case study from the Perception task.}
    \label{fig:case-study}
\end{figure}

%% file: prompts/tasks/perception.tex
\begin{figure*}[h]
  \centering
\begin{mybox}[Perception Task Description]
\begin{obeylines}
Perception tasks focus on observing and describing what is directly visible in a time series. Models are expected to identify surface-level patterns such as trends, periodicity, and anomalies, requiring strong observational and descriptive abilities.
\end{obeylines}
\end{mybox}
\caption{Description of the Perception task.}
\label{fig:desc-task-perception}
\end{figure*}

%% file: prompts/tasks/reasoning.tex
\begin{figure*}[h]
  \centering
\begin{mybox}[Reasoning Task Description]
\begin{obeylines}
Reasoning tasks require models to go beyond observation and construct logical explanations or conclusions from time series data. This category broadly covers inferring causes, discovering relationships, applying rules, and performing quantitative analysis, demanding strong analytical and multi-step thinking capabilities.
\end{obeylines}
\end{mybox}
\caption{Description of the Reasoning task.}
\label{fig:desc-task-reasoning}
\end{figure*}

%% file: prompts/tasks/prediction.tex
\begin{figure*}[h]
  \centering
\begin{mybox}[Prediction Task Description]
\begin{obeylines}
Prediction tasks ask models to forecast future values or outcomes based on historical time series. Models need to capture temporal dependencies and extrapolate patterns, requiring both quantitative estimation and event anticipation skills.
\end{obeylines}
\end{mybox}
\caption{Description of the Prediction task.}
\label{fig:desc-task-prediction}
\end{figure*}

%% file: prompts/tasks/decision_making.tex
\begin{figure*}[h]
  \centering
\begin{mybox}[Decision Making Task Description]
\begin{obeylines}
Decision-making tasks require models to produce actionable judgments informed by time series analysis. This encompasses both qualitative domain assessments and quantitative decision strategies, calling for the integration of analytical reasoning with domain-specific knowledge.
\end{obeylines}
\end{mybox}
\caption{Description of the Decision-Making task.}
\label{fig:desc-task-decision-making}
\end{figure*}

%% file: prompts/modalities/text.tex
\begin{figure*}[h]
  \centering
\begin{mybox}[Text Modality Description]
\begin{obeylines}
Text modality encodes time series as plain numerical sequences. It emphasizes precise arithmetic and statistical reasoning over raw numeric values, without relying on visual or spatial perception.
\end{obeylines}
\end{mybox}
\caption{Description of the Text modality.}
\label{fig:desc-modality-text}
\end{figure*}

%% file: prompts/modalities/visual.tex
\begin{figure*}[h]
  \centering
\begin{mybox}[Visual Modality Description]
\begin{obeylines}
Visual modality represents time series as rendered line chart images. It emphasizes spatial perception and visual pattern recognition, requiring models to interpret curves, axes, and chart structures without access to raw numbers.
\end{obeylines}
\end{mybox}
\caption{Description of the Visual modality.}
\label{fig:desc-modality-visual}
\end{figure*}

%% file: prompts/modalities/mix.tex
\begin{figure*}[h]
  \centering
\begin{mybox}[Both Modality Description]
\begin{obeylines}
Mix modality provides both numerical text sequences and chart images simultaneously. It enables cross-modal reasoning by combining token-level arithmetic with pixel-level visual interpretation, allowing models to leverage complementary information from both formats.
\end{obeylines}
\end{mybox}
\caption{Description of using both textual and visual modality.}
\label{fig:desc-modality-mix}
\end{figure*}

%% file: prompts/models/qwen3-8b.tex
\begin{figure*}[h]
  \centering
\begin{mybox}[Qwen3-8B Description]
\begin{obeylines}
A small text-only language model with 8 billion parameters from Alibaba. It offers fast inference at low cost (input: \$0.12/M tokens, output: \$0.45/M tokens), making it suitable for tasks that require quick, cost-efficient responses. Its compact scale trades off depth of reasoning for speed and affordability.
\end{obeylines}
\end{mybox}
\vspace{-12pt}
\caption{Qwen3-8B description.}
\label{fig:desc-model-qwen3-8b}
\end{figure*}

%% file: prompts/models/qwen3-32b.tex
\begin{figure*}[h]
  \centering
\begin{mybox}[Qwen3-32B Description]
\begin{obeylines}
A medium-sized text-only language model with 32 billion parameters from Alibaba. It provides a stronger capacity for analytical and mathematical reasoning than smaller models, at a moderate cost (input: \$0.16/M tokens, output: \$0.64/M tokens). It balances capability and efficiency for tasks requiring careful quantitative analysis.
\end{obeylines}
\end{mybox}
\caption{Qwen3-32B description.}
\label{fig:desc-model-qwen3-32b}
\end{figure*}

%% file: prompts/models/llama-3.3-70b-instruct-turbo.tex
\begin{figure*}[h]
  \centering
\begin{mybox}[LLaMA-3.3-70B-Turbo Description]
\begin{obeylines}
A large text-only language model with 70 billion parameters from Meta. Its scale provides advanced capabilities for complex multi-step logical inference and nuanced language understanding, at a higher cost (input: \$0.25/M tokens, output: \$0.75/M tokens). It is best suited for demanding tasks where smaller models fall short.
\end{obeylines}
\end{mybox}
\caption{LLaMA-3.3-70B-Turbo description.}
\label{fig:desc-model-llama-3-3-70b-instruct-turbo}
\end{figure*}

%% file: prompts/models/qwen3-vl-8b-instruct.tex
\begin{figure*}[h]
  \centering
\begin{mybox}[Qwen3-VL-8B-Instruct Description]
\begin{obeylines}
A small vision-language model with 8 billion parameters from Alibaba, supporting both visual and multimodal inputs. It offers cost-effective visual understanding at low cost (input: \$0.12/M tokens, output: \$0.45/M tokens), suitable for straightforward visual perception tasks where computational budget is limited.
\end{obeylines}
\end{mybox}
\caption{Qwen3-VL-8B-Instruct description.}
\label{fig:desc-model-qwen3-vl-8b-instruct}
\end{figure*}

%% file: prompts/models/qwen3-vl-32b-instruct.tex
\begin{figure*}[h]
  \centering
\begin{mybox}[Qwen3-VL-32B-Instruct Description]
\begin{obeylines}
A powerful vision-language model with 32 billion parameters from Alibaba, supporting both visual and multimodal inputs. It combines strong visual perception with solid analytical reasoning capabilities, at a moderate cost (input: \$0.16/M tokens, output: \$0.64/M tokens). It is well-suited for tasks requiring sophisticated interpretation of visual inputs.
\end{obeylines}
\end{mybox}
\caption{Qwen3-VL-32B-Instruct description.}
\label{fig:desc-model-qwen3-vl-32b-instruct}
\end{figure*}

%% file: prompts/models/glm-4.5v.tex
\begin{figure*}[h]
  \centering
\begin{mybox}[GLM-4.5V Description]
\begin{obeylines}
A large mixture-of-experts vision-language model with 106 billion parameters from Zhipu AI, supporting both visual and multimodal inputs. Its MoE architecture enables efficient scaling with strong visual and domain-specific capabilities, at a higher cost (input: \$0.20/M tokens, output: \$1.10/M tokens). It is designed for tasks demanding nuanced visual judgment and quantitative analysis.
\end{obeylines}
\end{mybox}
\caption{GLM-4.5V description.}
\label{fig:desc-model-glm-4-5v}
\end{figure*}

%% file: prompts/models/qwen3.5-397b-a17b.tex
\begin{figure*}[h]
  \centering
\begin{mybox}[Qwen3.5-397B-A17B Description]
\begin{obeylines}
A very large mixture-of-experts text-only language model with 397 billion parameters from Alibaba, with 397 billion total parameters and 17 billion active parameters per forward pass. Its MoE architecture delivers strong reasoning capabilities at a relatively moderate active compute cost, though pricing reflects its scale (input: \$0.65/M tokens, output: \$3.60/M tokens). It is designed for complex analytical tasks requiring deep multi-step thinking.
\end{obeylines}
\end{mybox}
\caption{Qwen3.5-397B-A17B description.}
\label{fig:desc-model-qwen3-5-397b-a17b}
\end{figure*}

%% file: prompts/models/kimi-k2.5.tex
\begin{figure*}[h]
  \centering
\begin{mybox}[Kimi-K2.5 Description]
\begin{obeylines}
A large vision-language model from Moonshot AI with built-in chain-of-thought reasoning capabilities, supporting both visual and multimodal inputs. It combines strong visual perception with multi-step logical reasoning, at a moderate cost (input: \$0.5/M tokens, output: \$2.80/M tokens). It is suited for tasks requiring integrated visual understanding and structured inference.
\end{obeylines}
\end{mybox}
\caption{Kimi-K2.5 description.}
\label{fig:desc-model-kimi-k2-5}
\end{figure*}